\documentclass[preprint,journal]{vgtc}       

\ifpdf
  \pdfoutput=1\relax                   
  \usepackage{graphicx}                
  \DeclareGraphicsExtensions{.pdf,.png,.jpg,.jpeg} 
\else
  \usepackage{graphicx}                
  \DeclareGraphicsExtensions{.eps}     
\fi%

\graphicspath{{figures/}{pictures/}{images/}{./}} 

\usepackage{microtype}                 
\PassOptionsToPackage{warn}{textcomp}  
\usepackage{textcomp}                  
\usepackage{times}                     
\usepackage{cite}                      
\usepackage{tabu}                      
\usepackage{booktabs}                  

\onlineid{1244}

\vgtccategory{Research}
\vgtcpapertype{algorithm/technique}

\usepackage{float}
\usepackage{longtable,booktabs}
\usepackage{algorithm}
\usepackage{multirow}
\usepackage{amsmath}
\usepackage{amsfonts}
\usepackage[pagebackref=true,breaklinks=true,letterpaper=true,colorlinks,bookmarks=false]{hyperref}

\newcommand{\mbf}[1]{\mathbf{#1}}
\DeclareMathOperator*{\argmin}{argmin}
\newcommand{\T}{\mathcal{T}}
\newcommand{\F}{\mathcal{F}}
\newcommand{\s}{\mbf{s}}
\newcommand{\Ip}{\mbf{I}_\text{p}}
\newcommand{\Ic}{\mbf{I}_\text{c}}
\newcommand{\xc}{\mbf{x}_\text{c}}
\newcommand{\xs}{\mbf{x}_\text{s}}
\newcommand{\dc}{\mbf{d}_\text{c}}

\def\eg{\emph{e.g}.} 
\def\ie{\emph{i.e}.} 
 
\def\etc{\emph{etc}.} 
 
\def\etal{\emph{et al}.}

\newcommand{\abs}[1]{ \left\lvert#1\right\rvert} 
\newcommand{\norm}[1]{\left\lVert#1\right\rVert} 

\makeatletter
\newcommand{\thickhline}{%
	\noalign {\ifnum 0=`}\fi \hrule height 1pt
	\futurelet \reserved@a \@xhline
}
\makeatother

\title{DeProCams: Simultaneous Relighting, Compensation and Shape Reconstruction for Projector-Camera Systems}

\author{Bingyao Huang,~ Haibin Ling}
\authorfooter{
\item Bingyao Huang and Haibin Ling are with the Department of Computer Science, Stony Brook University, Stony Brook, NY 11794 USA. \\E-mail: \{bihuang, hling\}@cs.stonybrook.edu.
}

\shortauthortitle{Huang \MakeLowercase{\textit{et al.}}: DeProCams: Simultaneous Relighting, Compensation and Shape Reconstruction for Projector-Camera Systems}

\abstract{
  Image-based relighting, projector compensation and depth/normal reconstruction are three important tasks of projector-camera systems (ProCams) and spatial augmented reality (SAR). Although they share a similar pipeline of finding projector-camera image mappings, in tradition, they are addressed independently, sometimes with different prerequisites, devices and sampling images. In practice, this may be cumbersome for SAR applications to address them one-by-one. In this paper, we propose a novel end-to-end trainable model named DeProCams to explicitly learn the photometric and geometric mappings of ProCams, and once trained, DeProCams can be applied simultaneously to the three tasks. DeProCams explicitly decomposes the projector-camera image mappings into three subprocesses: shading attributes estimation, rough direct light estimation and photorealistic neural rendering. A particular challenge addressed by DeProCams is occlusion, for which we exploit epipolar constraint and propose a novel differentiable projector direct light mask. Thus, it can be learned end-to-end along with the other modules. Afterwards, to improve convergence, we apply photometric and geometric constraints such that the intermediate results are plausible. In our experiments, DeProCams shows clear advantages over previous arts with promising quality and meanwhile being fully differentiable. Moreover, by solving the three tasks in a unified model, DeProCams waives the need for additional optical devices, radiometric calibrations and structured light.
}


\teaser{
  \centering
  \includegraphics[width=1\linewidth]{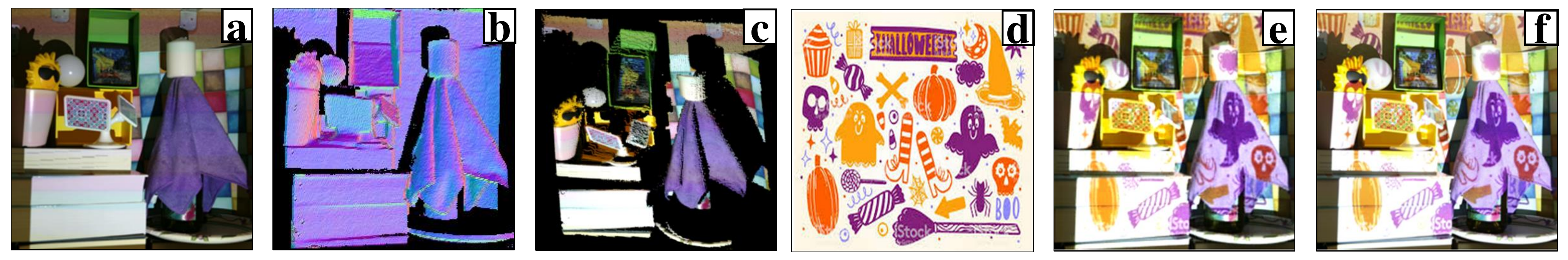}
  \caption{Deep Projector-Camera systems (DeProCams): 
      \textbf{(a)} scene under plain white illumination,
      \textbf{(b)} estimated normal map,  
      \textbf{(c)} estimated colored point cloud viewed from a different pose,
      \textbf{(d)} a new projector input pattern,
      \textbf{(e)} inferred relit result, and 
      \textbf{(f)} camera-captured ground truth, \ie, (d) projected to (a). }\label{fig:teaser}
}

\vgtcinsertpkg

\begin{document}


\firstsection{Introduction}\label{sec:intro}

\maketitle

Projector-camera systems (ProCams) have many applications in projection mapping/spatial augmented reality (SAR) \cite{raskar2001shader, narita2017dynamic, siegl2015real, sueishi2015robust, grundhofer2018recent, ueda2020illuminated, takatori2019large, iwai2018non, takezawa2019material}, which usually involves three important tasks: image-based relighting \cite{ren2015image, sen2005dual, han2014fast, masselus2003relighting, oya2017image, peers2009compressive, guo2019relightables}, projector compensation \cite{raskar2001self, raskar2003ilamps, tardif2003multi, boroomand2016saliency, narita2017dynamic, nayar2003projection, grossberg2004making, yoshida2003virtual, ashdown2006robust, aliaga2012fast, bimber2008visual,grundhofer2015robust,huang2019compennet,huang2019compennet++, kageyama2020prodebnet}, and structured light shape reconstruction \cite{Geng2011, moreno2012simple, o20163d, vo2015texture}. An example is shown in \autoref{fig:teaser}, where image-based relighting and depth/normal estimation are performed using the proposed DeProCams.

Image-based relighting aims to simulate the camera-captured image when a novel projector image is projected to the scene. It can synthesize realistic projection mapping effects without actual project-and-capture operations, which can be useful for SAR testing/debugging and virtual storing and sharing SAR setups.
A typical solution for image-based relighting is light transport matrix (LTM) \cite{zongker1999environment, debevec2000acquiring, sen2005dual, o20163d, chiba2018ultra, wang2009kernel, o2010optical}, which starts by projecting and capturing various sampling images and then estimate an LTM from the collected images, finally a camera-captured image can be inferred by applying the LTM to a novel projector input image. Although LTM produces accurate relighting results, it may be time-consuming \cite{sen2005dual}, or requires additional optical devices such as lasers and a second camera \cite{wang2009kernel} and beamsplitters \cite{o2010optical}. Moreover, projector-camera radiometric calibration \cite{grossberg2004modeling, debevec2008recovering} is required and thus may be inconvenient for some projection mapping/SAR applications that frequently adjust the projector settings.

Projector compensation aims to modify a projector image such that when projected to the scene it can compensate for the 
disturbance from the scene geometry and photometry, and create desired viewer-perceived effects. This technique is often applied in SAR applications, \eg, appearance editing \cite{aliaga2012fast, siegl2015real} and projection-based transparent effects \cite{iwai2011document, yoshida2008transparent}. A typical pipeline \cite{nayar2003projection, grundhofer2015robust, bimber2008visual} is to first find pixel geometric mappings using structured light, then estimates pixel-wise photometric transfer functions from the geometrically registered sampling images, finally a projector input image can be compensated by applying the inverse photometric transfer functions. 

Shape reconstruction aims to estimate the scene depth and normal using a different set of sampling images other than the image-based relighting and the projector compensation ones. Knowing the scene shape can significantly improve the accuracy of projected augmentations, and aid depth-aware interactive SAR applications \cite{benko2009beyond, hochreiter2015touch}. A typical shape reconstruction method is structure light \cite{Geng2011} that projects and captures hand-crafted patterns, then estimates depth (and normal) using the matched pattern pixels and geometric calibrations.

Because the three tasks have different prerequisites, optical devices and sampling images, some SAR applications solve them independently, although they share a similar pipeline of projecting and capturing sampling images and then finding projector-camera image mappings.

In this paper, we show that it is possible to solve the three tasks simultaneously without additional optical devices, radiometric calibrations or different sampling images (\eg, structured light patterns). In particular, we explicitly model projector-camera image mappings as a geometry- and attributes-aware shading problem and solve it using an end-to-end trainable model named \textbf{DeProCams} (Deep Projector-Camera systems), which consists of two modules: \textit{DepthToAttribute} and \textit{ShadingNet}.

From a learnable depth map, the \textbf{DepthToAttribute} module explicitly calculates the shading attributes such as projector direct light rays, surface normal, view and reflection directions, \etc\ To address the \textit{occlusion} challenge in shading, a novel \textit{differentiable} projector direct light mask is derived by exploring the related epipolar geometry, and can be efficiently computed using differentiable image warping \cite{jaderberg2015spatial}, tensor sorting and image gradient. Then, initial rough direct light components are estimated to constrain photometry and geometry and to provide a good initialization for ShadingNet's photorealistic direct and indirect light rendering. 

Afterwards, the \textbf{ShadingNet} module is designed to aggregate intermediate shading attributes and rough direct light, and to render the camera-captured image with photorealistic direct and indirect light. Moreover, to improve model robustness and convergence, we incorporate multiple photometric and geometric constraints, such as projector direct light mask consistency, rough diffuse shading consistency and depth/normal/pixel mapping smoothness. Because of the explicit modeling of light-geometry-material interactions and sufficient constraints, our depth map is learned from the sampling images without its ground truth.

Finally, after a few minutes of data capturing and training, three SAR tasks, \ie, \textit{image-based relighting}, \textit{projector compensation} and \textit{depth/normal estimation} can be simultaneously performed by the proposed DeProCams. 

Our contributions can be summarized as follows:
\begin{itemize}
  \item The proposed DeProCams is, to our best knowledge, the first to simultaneously perform three SAR tasks with one learned model: image-based relighting, projector compensation and depth/normal estimation. 
  \item The proposed DeProCams does not require additional optical devices, radiometric calibrations and structured light patterns.
  \item Multiple projector-camera photometric and geometric constraints are incorporated into DeProCams to improve model convergence and performance, and we propose a \textit{differentiable} projector direct light mask to explicitly address the occlusion issue. These techniques are expected to facilitate future works in deep learning-based SAR application.
  \item We construct a simultaneous image-based relighting and shape reconstruction benchmark for projector-camera systems, which allows future works to compare with DeProCams using our training and testing dataset.
\end{itemize}

For the benefit of the society, the source code, training and evaluation benchmark dataset and experimental results are made publicly available at \url{https://github.com/BingyaoHuang/DeProCams}.

\section{Related Work}\label{sec:related_work}

\subsection{Image-based relighting}
In tradition, image-based relighting methods rely on estimating a light transport matrix (LTM) \cite{zongker1999environment,debevec2000acquiring, sen2005dual, o20163d, chiba2018ultra, wang2009kernel, ren2015image} from projected and captured sampling image pairs, such that each camera pixel's irradiance is modeled as a linear combination of \textit{all} projector pixels' radiances.

Early methods \cite{debevec2000acquiring, wenger2005performance} directly estimate LTM using brute force sampling that requires plenty of HDR images, memory and time. To improve LTM estimation efficiency, later methods exploit the sparsity and coherence of LTM \cite{zongker1999environment, ng2003all, masselus2003relighting, sen2005dual, peers2009compressive, ren2015image, wang2009kernel, chiba2018ultra}, leading to reduced sampling images and computation. 
For example, \cite{sen2005dual} leverage adaptive sampling to reduce LTM reconstruction time to two hours. With additional RGB lasers and a second camera, Wang \etal \cite{wang2009kernel} use kernel Nystr{\"o}m to approximate LTM. Ren \etal  \cite{ren2015image} model each LTM entry as a nonlinear function of pixels and light locations, then use multilayer neural networks to approximate it.	Once reconstructed, LTM can be applied to a novel projector input image to simulate the relit scene. 

Typical LTM-based relighting methods focus on accurate indirect light estimation and usually a camera cannot directly see projected patterns \cite{wang2009kernel}. While other SAR applications focus on direct light such as structured light \cite{Geng2011}, projection-based appearance editing \cite{raskar2001shader, grossberg2004making, siegl2015real} and augmentation \cite{iwai2018non, iwai2011document, narita2017dynamic},
where camera directly sees projected patterns (\autoref{fig:coord}), \ie, \emph{projector direct light dominates}. For those applications, estimating a full LTM is time consuming and it may be simplified. For example, some approaches \cite{nayar2003projection, grossberg2004making, grundhofer2015robust} assume that each camera pixel is only coupled with a \textit{single} projector pixel that contributes most of the irradiance. Then projector-camera pixel-wise mappings can be applied to efficiently solve \textit{projector compensation} and \textit{attributes from shading}. Following this kind of work in SAR applications, in this paper, we focus on a single projector-camera pair (\autoref{fig:coord}) where projector direct light dominates, but unlike \cite{nayar2003projection, grossberg2004making, grundhofer2015robust}, we do not assume pixel-wise mapping and thus can estimate indirect light.

\subsection{Projector compensation} 
Projector compensation estimates a projector input image given its corresponding desired camera-captured image and can be roughly categorized into full compensation, \ie, both geometric and photometric compensation and partial ones.	

Wetzstein \etal \cite{wetzstein2007radiometric} use an inverse LTM to obtain accurate indicate light compensation, \eg, glasses, and thus projector-camera radiometric calibration is required as a prerequisite of LMT. In practice, the camera radiometric calibration is simple but the projector calibration is nontrivial. Moreover, recalibration is needed when the projector settings change, such as brightness, contrast and color profiles. Thus it may be inconvenient for some SAR applications that frequently adjust projector settings. To waive radiometric calibrations,  Grundh{\"o}fer and Iwai \cite{grundhofer2015robust} encode projector-camera radiometric responses and photometric transfer function using a pixel-wise thin-plate-spine (TPS), such that the composite transfer function can be estimated using RGB sampling images. 
To further waive the structured light patterns, Huang and Ling \cite{huang2019compennet++} propose a full compensation method that jointly learns 2D geometry and photometry from the sampling images using a deep neural network. More detailed reviews can be found in \cite{bimber2008visual, grundhofer2018recent}, and some recent representative works can be found in \cite{grundhofer2015robust} and \cite{huang2019compennet, huang2019compennet++, kageyama2020prodebnet,huang2020end}.

\subsection{Attributes from shading} 
Attributes from shading \cite{barron2014shape} aims to estimate scene attributes from camera-captured images, such as surface normal, reflectance, light directions and depth, \etc\ To avoid complex indirect light modeling, some simplifications are imposed, such as assuming distant light in photometric stereo (PS) or direct light dominance in structured light (SL), respectively. 
In the following we briefly review studies in PS and SL that relate most to our work. 
The two differences between PS and SL are (1) PS assumes light sources with fixed radiances and varying directions placed at infinity while SL assumes a projector with varying output patterns placed at the fixed near field. (2) PS estimates depth and normal from light and surface interactions/shadings, while SL computes depth using 2-D pixel mappings and triangulation. 

\noindent\textbf{Photometric stereo} (PS) \cite{woodham1980photometric, chen2019self, shi2019benchmark, santo2017deep, taniai2018neural} estimates surface normal by varying the light directions while fixing the view direction. In its earliest form \cite{woodham1980photometric} the surface is assumed Lambertian and the surface normal is obtained by solving linear equations. More complex BRDFs are considered later~\cite{georghiades2003incorporating, shi2013bi}. Recently, Santo \etal \cite{santo2017deep} leverage deep neural networks to solve this problem using supervised training data. To avoid creating normal ground truth, Taniai \etal \cite{taniai2018neural} propose a self-supervised photometric stereo network that uses image reconstruction loss. To deal with uncalibrated cases, Chen \etal \cite{chen2019self} propose a self-calibrating method to jointly estimate unknown light directions and surface normal from non-Lambertian surfaces. 

\noindent\textbf{Structured light} (SL) \cite{Geng2011, moreno2012simple, o20163d, vo2015texture, li2019pro} estimates scene depth by matching projected and captured hand-crafted patterns, then matched 2-D pixel coordinates are triangulated \cite{hartley2003multiple} to recover the 3-D point cloud. However, this two-step strategy heavily depends on projector direct light and thus 2-D pixel matching may fail in strong indirect light regions \cite{o20163d, chiba2018ultra, huang2019compennet++, nayar2006fast}. To address this issue, O'Toole \etal \cite{o20163d} combine LTM with SL to jointly estimate scene depth, albedo and indirect light.
Although PS and SL can recover the geometry of the scene, they alone are unable to solve  image-based relighting or projector compensation. 

\subsection{Relationship to DeProCams}
Recently, deep learning-based solutions have been proposed for related works, such as deep shading/neural rendering \cite{hermosilla2019deep, nalbach2017deep, li2018differentiable}, projector compensation \cite{huang2019compennet, huang2019compennet++, kageyama2020prodebnet} and attributes from shading \cite{deschaintre2018single, li2017modeling, qi2018geonet,li2018learning, yu2019inverserendernet}. However, these methods either require scene modeling or have simple assumptions on scene geometry, or need ground truth attributes supervision.

The question is, without scene modeling or ground truth attributes, can we learn a model from projector-camera sampling images, and apply it \textit{simultaneously} to three tasks, \ie, image-based relighting, projector compensation and depth/normal reconstruction? It is shown in CompenNet++~\cite{huang2019compennet++} to be possible for projector compensation, but restricted to smooth occlusion-free surfaces. Moreover, the depth and normal involved in the projector-camera system are ignored. Both limitations make it unable to perform image-based relighting in general scenes with occlusions or depth/normal reconstruction. Thus, directly applying CNNs to projector-camera systems is hard without modeling the light-geometry-material interactions.
By contrast, our DeProCams explicitly models light and geometry (depth/normal) interactions, and leverages additional photometric and geometric constraints, meanwhile performing three tasks simultaneously.

\section{Deep Projector-Camera Systems}\label{sec:problem}
\subsection{Problem formulation}\label{subsec:problem_formulation}
\begin{figure}[!t]
  \includegraphics[width=1\linewidth]{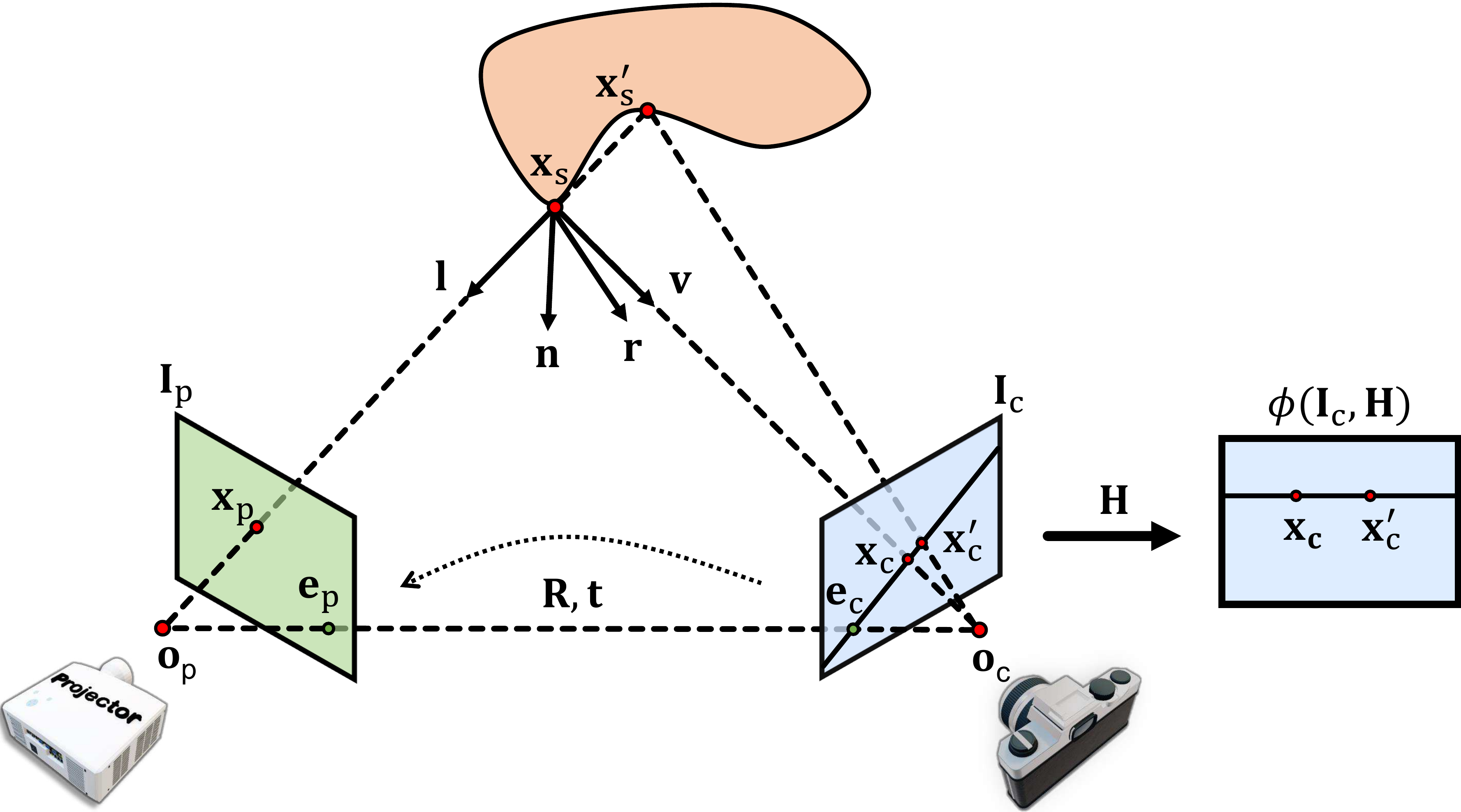}
  \caption{\textbf{Coordinate system of our projector-camera setup}. With the geometric calibration, a rectification transformation $\mbf{H}$ can be calculated to transform the camera-captured image such that all epipolar lines are parallel. }   \label{fig:coord}
\end{figure}

As shown in \autoref{fig:coord}, we model the projector-camera image formation as a mapping from a projector input RGB image $ \Ip $ to a camera-captured RGB image $ \Ic$:
\begin{equation}\label{eq:render_noncalib}
  \Ic = \pi(\Ip)
\end{equation}
where $ \pi $ is the composite image transfer function that contains nonlinear projector and camera radiometric responses, image projection functions and projector light ray, scene geometry and material interactions, \etc\ Apparently \autoref{eq:render_noncalib} has no closed form solution and one intuition is to directly learn $ \pi $ from sampling image pairs like $ \Ic$ and $ \Ip$ using a general image-to-image translation CNN, \eg, U-Net \cite{ronneberger2015u} or Pix2pix \cite{isola2017image}. It may work if the training set is large or data augmentation is available. However, SAR applications require online data capturing and thus it is impractical to capture a large sampling set. Moreover, data augmentation such as rotation and random cropping cannot be applied due to strong couplings between projector-camera pixels. As a result, only a small number of sampling images are available for model training. To address this issue, we leverage domain knowledge of projector-camera image formation, such as light-geometry-material interactions and photometric and geometric constraints. 

Note that our method is inspired by CompenNet++ \cite{huang2019compennet++}, which disentangles 2D pixel geometric mappings and photometry from projector-camera systems. Compared with \cite{huang2019compennet++}, our DeProCams explicitly separates shading attributes and depth from the 2D pixel geometric mappings; and we explicitly compute rough shadings before applying photometric neural rendering. More specifically, we decompose $\pi$ into a geometry-aware module \textbf{DepthToAttribute} $\T$ to compute rough shading attributes and a photometry module \textbf{ShadingNet} $\F$ for photorealistic neural rendering. In particular, DepthToAttribute contains a learnable depth map (in the canonical camera frontal view) $\dc \in \mathbb{R}^{\text{h}\times\text{w}\times 1}$ and ShadingNet $\F$ is a neural network that consists of learnable convolutional filters. The learnable parameters of DeProCams are $ \{\dc, \F\}_\theta$. Note that $\dc$ and $ \F $ are jointly optimized during DeProCams training, thus the term ``learnable".

\subsection{DepthToAttribute $\T$}\label{subsec:depth_to_attribute}
\subsubsection{Shading attributes}\label{sec:attributes}
As shown in \autoref{fig:coord}, we assume that the world origin is at the camera optical center $ \mbf{o}_\text{c} $, and denote the camera and the projector intrinsics as $ \mbf{K}_\text{c} $ and $ \mbf{K}_\text{p} $, and denote their relative rotation and translation as $ \mbf{R} $ and $ \mbf{t} $, respectively. For a surface point $ \xs\in \mathbb{R}^{3} $ in the scene, let $ \xc, \mbf{x}_\text{p} \in \mathbb{R}^{2} $ be its pixel coordinates in the camera image $ \Ic $ and the projector image $ \Ip $, respectively. Then the depth of $ \xs $ is given by $ \mathrm{d}_\text{c} = \dc(\xc) $. The geometric relationships between $ \xs $, $ \xc $ and $ \mbf{x}_\text{p} $ are given by:
\begin{equation}\label{eq:coord_map}
\xs = \mbf{K}_\text{c}^{-1}\bar{\mbf{x}}_\text{c}\mathrm{d}_\text{c}, \quad \quad
\bar{\mbf{x}}_\text{p} = \mbf{K}_\text{p}[\mbf{R}|\mbf{t}]\bar{\mbf{x}}_\text{s}
\end{equation}
where $ \bar{\mbf{x}} $ is the homogeneous coordinate of $ \mbf{x} $. Given the 3-D coordinates $\xs$ in \autoref{eq:coord_map}, shading attributes such as the projector direct light ray direction $ \mbf{l} $, the reflection ray direction $ \mbf{r} $ and the camera view direction $ \mbf{v} $ are computed by:
\begin{align}
&\mbf{n} = \partial_x\mbf{\xs}\times\partial_y\xs  &\ &\mbf{l} = \mbf{o}_\text{p} - \xs \label{eq:normal}\nonumber\\ 
&\mbf{r} = 2(\mbf{n}\cdot\mbf{l})\mbf{n} - \mbf{l} &\ &\mbf{v} = \mbf{o}_\text{c} - \xs
\end{align}
where $ \times $ is vector cross product. Note that in above equation $ \mbf{n}, \mbf{l}, \mbf{r}, \mbf{v} \in \mathbb{R}^3 $ are shadings attributes of a single surface point $ \xs $. To extend \autoref{eq:normal} to all surface points, we abuse the notation and refer them as $ \mathbb{R}^{\text{h}\times\text{w}\times 3} $ maps in the following text.
Meanwhile, given the projector and camera pixel geometric mappings in \autoref{eq:coord_map}, we construct an image sampling/warping grid $\mbf{\Omega} \in \mathbb{R}^{\text{h}\times\text{w}\times 2}$ such that $ \mbf{x}_\text{p} =\mbf{\Omega}(\xc) $. Therefore, the projector input image warped to the canonical camera frontal view is given by:
\begin{equation}\label{eq:warp}	
  \mbf{I}'_\text{p} = \phi(\Ip, \mbf{\Omega})\odot\mbf{M}
\end{equation}
where $ \phi $ is a differentiable image interpolator \cite{jaderberg2015spatial}; $ \odot $ is element-wise multiplication, and $ \mbf{M} \in \mathbb{R}^{\text{h}\times\text{w}\times 1}$ is a mask and more details are given in \autoref{sec:mask}.

\subsubsection{Rough shadings}\label{sec:rough}
Afterwards, we can compute light-geometry-material interactions using a rough direct light shading model (\autoref{eq:rough}) under the direct light dominance assumption of SAR applications. \emph{It is worth noting that our rough shading model is inspired but different from the Phong model \cite{phong1975illumination, blinn1977models, shafer1985using} (thus the name ``rough'')}. Rather than using the actual diffuse and specular reflectance maps and shininess of Phong model \cite{phong1975illumination, blinn1977models, shafer1985using}, we use a rough approximation $\s$ instead and set shininess $ \alpha=1 $, because (1) estimating actual reflectance maps are difficult and unnecessary for our tasks; and (2) even with accurate reflectance maps, the empirical model (\autoref{eq:rough}) alone cannot produce photorealistic rendering. Therefore, \emph{the rough shadings only provide a simple yet plausible initial point for ShadingNet (\autoref{eq:photometry}), meanwhile, leverage photometric and geometric constraints (\autoref{sec:constraints})}. 
\begin{equation}\label{eq:rough}
  \mbf{I}^{\text{abnt}}_\text{c} = \s,\quad
  \mbf{I}^{\text{diff}}_\text{c} = \mbf{I}'_\text{p}\odot\s\odot(\mbf{n}\cdot \mbf{l}),\quad
  \mbf{I}^{\text{spec}}_\text{c} = \mbf{I}'_\text{p}\odot\dot{\s}\odot(\mbf{r}\cdot \mbf{v})^{\alpha}
\end{equation}
where 
$\s $ models rough ambient and surface reflectance, and is approximated by a scene image captured under plain white illumination (\autoref{fig:net}); $ \odot $ and $ \cdot $ are element-wise multiplication and dot product, respectively; and $ \dot{\s} $ is the grayscale image of $ \s $. 

\begin{figure*}[!t]
  \begin{center}
    \includegraphics[width=1\linewidth]{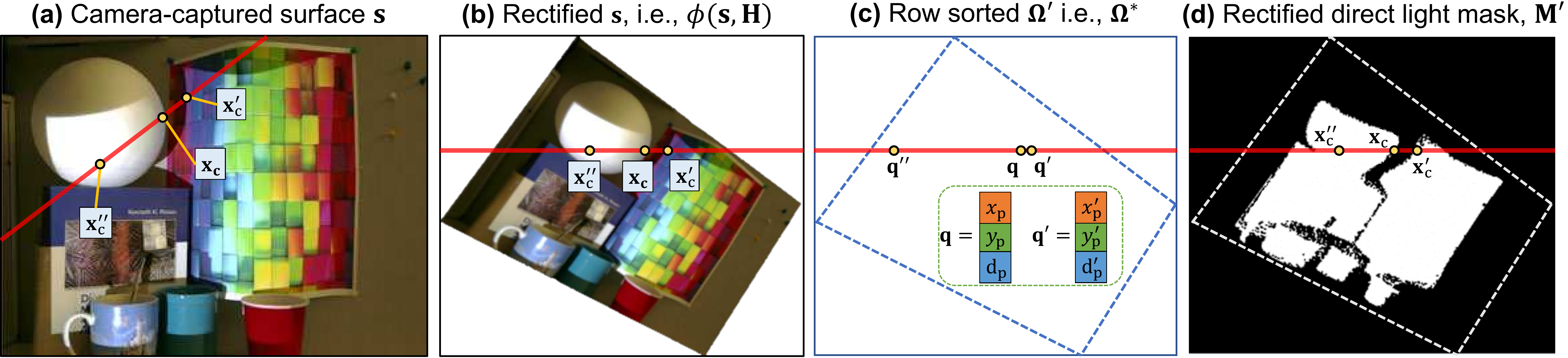}
    \caption{\textbf{Proposed differentiable projector direct light mask}. \textbf{(a)} Three points are on the same red epipolar line and finding occlusion requires 2-D searching. \textbf{(b)} Instead, we perform 1-D search in stereo rectified image. \textbf{(c)} To speed up pair-wise comparison, we row sort $ \mbf{\Omega}' $ in lexicographic order. After sorting,  $ \xc$'s and $ \mbf{x}'_\text{c}$'s corresponding points $ \mbf{q}, \mbf{q}' \in \mbf{\Omega}^{*} $ become neighbors, then we apply \autoref{eq:mask_grad} and find that  $ \mbf{x}'_\text{c} $ is occluded, since $ |x'_\text{p} - x_\text{p}| + |y'_\text{p} - y_\text{p}| < 1$ and $ |\mathrm{d}'_\text{p} - \mathrm{d}_\text{p}| > 0 $. All operations are differentiable and parallelizable, thus are efficient for training on GPU.	
    }   \label{fig:mask}
  \end{center}
\end{figure*}

\subsubsection{Differentiable projector direct light mask}\label{sec:mask}
\begin{figure*}[!t]
  \begin{center}
    \includegraphics[width=1\linewidth]{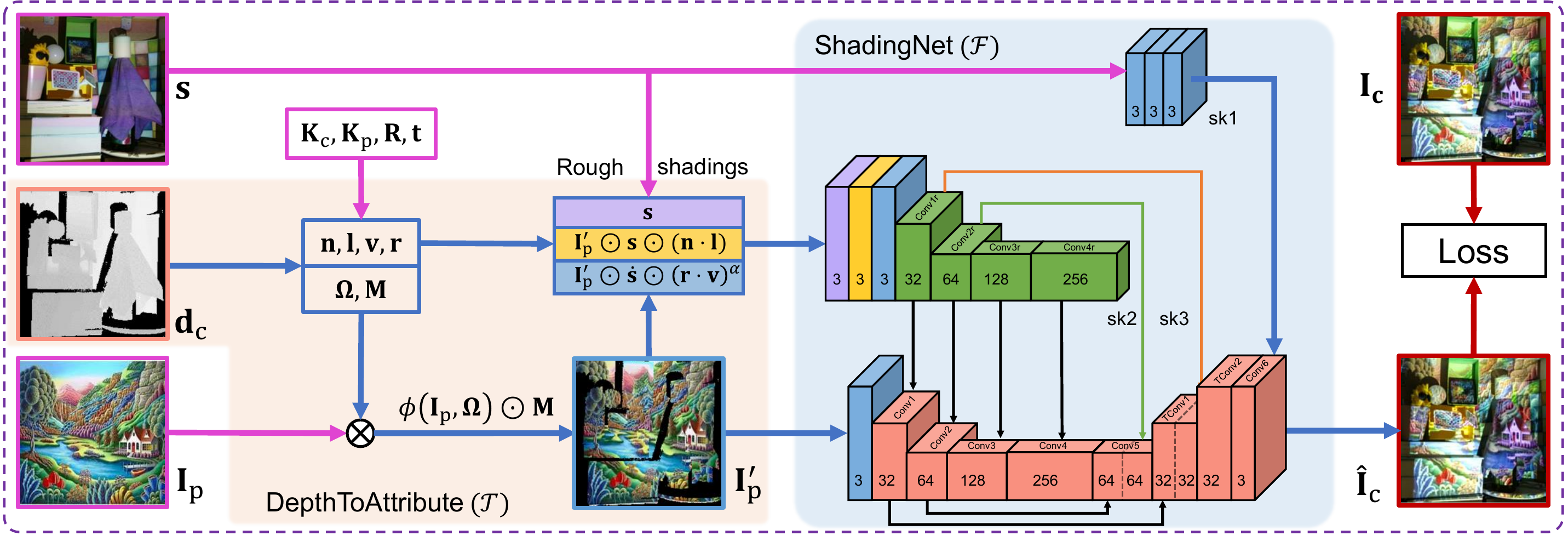}   
    \caption{\textbf{Deep projector-camera systems (DeProCams) architecture and training pipeline}. The inputs (magenta) are $\Ip$, $\s$, $ \mbf{K}_\text{c}, \mbf{K}_\text{p}, \mbf{R} $ and $ \mbf{t} $, and output is $\hat{\mbf{I}}_\text{c}$; { $\phi$} is an image warper. In the forward/inference stage, using the depth map $ \dc $  and calibration data $ \mbf{K}_\text{c}, \mbf{K}_\text{p}, \mbf{R} $ and $ \mbf{t} $, DepthToAttribute $ \T $ computes the shading attributes, \ie, $ \mbf{n}, \mbf{l}, \mbf{r}, \mbf{v}, \mbf{\Omega}, \mbf{M} $ and warps projector direct light image to the camera view as $ \mbf{I}'_\text{p}$.  Then, we compute three rough shadings.  Finally, ShadingNet $ \F $ gathers rough direct light shadings and $ \mbf{I}'_\text{p}$ to infer photorealistic camera-capture image $ \hat{\mbf{I}}_\text{c} $. During backpropagation, the loss (between $\hat{\mbf{I}}_\text{c}$ and $ \Ic $) gradient is backpropagated to update $ \dc $ and ShadingNet $ \F $ parameters jointly. }\label{fig:net} 
    \end{center}
\end{figure*}

Occlusion is a particular challenge to depth and normal reconstruction because it may make two camera pixels map to the same projector pixel (\autoref{fig:coord} and \autoref{fig:mask}), causing false projector direct light rays, while the depth-based image warping $\phi$ (\autoref{eq:warp} and \autoref{fig:net}) is unable to address it. Some previous methods \cite{wang2018occlusion, meister2018unflow, hur2017mirrorflow, godard2017unsupervised, garg2016unsupervised} leverage forward-backward consistency to find occlusions, assuming that only slight motion and lighting changes across views. However, for a projector-camera system such an assumption may be violated.
Therefore, we exploit epipolar geometry and propose a \textit{differentiable} projector direct light mask $\mbf{M}$ to address occlusions below.	

As shown in \autoref{fig:coord} and \autoref{fig:mask}, two surface points $ \xs $ and $ \mbf{x}'_\text{s} $ lie on the same projector ray, therefore their projections $ \xc $ and $ \mbf{x}'_\text{c} $ must lie on the same epipolar line. From \autoref{fig:coord} and \autoref{fig:mask}, we see that $ \mbf{x}'_\text{s} $ is occluded by $ \xs $, and thus only $ \xc $ is illuminated by the projector direct light. This type of occlusion can be modeled by a projector direct light mask $ \mbf{M} $, such that for any camera pixel $ \xc $ on the epipolar line $  \overrightarrow{\mbf{e}_\text{c}\xc}$ ($ \mbf{e}_\text{c} $ is the camera epipole), its pixel value in $ \mbf{M} $ is given by:
\begin{equation}\label{eq:mask}
\mbf{M}(\xc) =
\begin{cases}
1, & \forall \mbf{x}'_\text{c} \in \overrightarrow{\mbf{e}_\text{c}\xc},\ \mbf{x}_\text{p} \not\approx \mbf{x}'_\text{p} \  \text{OR}\  \mathrm{d}_\text{p} < \mathrm{d}'_\text{p} \\
0, & \text{otherwise}
\end{cases}
\end{equation}
where $ \mbf{x}_\text{p} =\mbf{\Omega}(\xc) $ and $ \mbf{x}'_\text{p} =\mbf{\Omega}(\mbf{x}'_\text{c}) $ are the corresponding projector pixel coordinates (\autoref{fig:coord}); $ \mathrm{d}_\text{p} = ([\mbf{R}|\mbf{t}]\bar{\mbf{x}}_\text{s})_z $  and $ \mathrm{d}'_\text{p} = ([\mbf{R}|\mbf{t}]\bar{\mbf{x}}'_\text{s})_z $ are the depths of $ \xc $ and $ \mbf{x}'_\text{c} $ in the projector view space, respectively. Then, we apply a stereo rectification \cite{hartley2003multiple} $ \mbf{H} $ to reduce the 2-D search to 1-D  (\autoref{fig:mask}(b)). 
However, even with 1-D search, the pair-wise comparison in \autoref{eq:mask} for each pixel on each epipolar line is still expensive. The logical OR operator in \autoref{eq:mask} implies that depth comparison is only needed when 
$ \xc $ and $ \mbf{x}'_\text{c} $ map to the same \textit{projector} pixel, \ie, $ \mbf{x}_\text{p} \approx \mbf{x}'_\text{p} $ (\autoref{fig:coord}). 
Therefore, for each \textit{rectified} epipolar line, if we sort each pixel's corresponding projector pixel coordinates $ \mbf{x}_\text{p} $ in lexicographical order, the pair-wise comparisons in \autoref{eq:mask} can be performed only in a small neighborhood, \eg, $ \pm1 $ pixel.
  
To make the above operations \textit{differentiable}, we concatenate the sampling/warping grid $\mbf{\Omega}$ and the projector-view depth map $ \mbf{d}_\text{p}$ along the channel and rectify them using the differentiable image interpolator \cite{jaderberg2015spatial} by $ \mbf{\Omega}' = \phi([\mbf{\Omega}, \mbf{d}_\text{p}], \mbf{H}),\  \mbf{\Omega}' \in \mathbb{R}^{\text{h}\times\text{w}\times 3} $. Next, we sort each $ 3\times1 $ element in each row of $ \mbf{\Omega}' $ (\ie, along epipolar line) and denote the sorted tensor $\mbf{\Omega}' $ as $\mbf{\Omega}^{*} $ (\autoref{fig:mask}(c)).
Afterwards, we compute the forward difference in the row direction, then the rectified direct light mask $ \mbf{M'} $ is given by:
\begin{equation}\label{eq:mask_grad}
\mbf{M'} =
\begin{cases}
1, & |\partial_x\mbf{\Omega}^{*}_{x}| + |\partial_x\mbf{\Omega}^{*}_{y}| > 1 \  \text{or}\   \partial_x\mbf{\Omega}^{*}_{z} < 0 \\
0, & \text{otherwise}
\end{cases}
\end{equation}
Note that the non-differentiable unit step function above is approximated by a differentiable clamp and ReLU \cite{nair2010rectified}: $ \min(\sigma*\text{ReLU}(\cdot),\ 1),\ \sigma \gg  1 $.
Then, we unrectify mask $ \mbf{M'} $ (\autoref{fig:mask}(d)) back to the camera view by $ \mbf{M} = \phi(\mbf{M'}, \mbf{H}^{-1}) $.
Finally, the projector direct light mask is applied to mask out false direct light pixels as shown in \autoref{fig:net}, and to enforce direct light mask consistency in \autoref{sec:mask_const}.

We implement the above geometric transformations using differentiable operations (Kornia \cite{eriba2019kornia}), and then gather them in \textbf{DepthToAttribute} $ \T $ (\autoref{fig:net}). It is worth noting that, unlike the WarpingNet in~\cite{huang2019compennet++} that only estimates a sampling/warping grid $ \mbf{\Omega} $, DepthToAttribute learns depth map and computes shading attributes and rough shadings:
\begin{equation}\label{eq:attributes}	
	\{\mbf{n}, \mbf{l}, \mbf{r}, \mbf{v}, \mbf{\Omega},  \mbf{M}, \mbf{I}'_\text{p}, \mbf{I}^{\text{diff}}_\text{c},\mbf{I}^{\text{spec}}_\text{c}\} = \T(\mbf{K}_\text{c}, \mbf{K}_\text{p}, \mbf{R}, \mbf{t})	
\end{equation}
Note that we omit $ \dc $ in above equation because it is inherently included in $ \T $. These attributes are then fed to ShadingNet to further infer photorealistic shadings and used as additional constraints (\autoref{sec:constraints}).

\subsection{ShadingNet}\label{subsec:shading_net}
To infer photorealistic direct and indirect light, we design an image-to-image translation CNN named \textbf{ShadingNet} $ \F $ to aggregate intermediate results and refine rough direct light shadings by:
\begin{equation}\label{eq:photometry}
  \hat{\mbf{I}}_\text{c} = \F(\mbf{I}'_\text{p},\ \mbf{I}^{\text{abnt}}_\text{c},\ \mbf{I}^{\text{diff}}_\text{c},\ \mbf{I}^{\text{spec}}_\text{c} )	
\end{equation} 
ShadingNet architecture is shown in \autoref{fig:net} and \autoref{tab:shadingnet_param}, and it takes four inputs: (1) rough ambient, \ie, the camera-captured surface image {$ \mbf{I}^{\text{abnt}}_\text{c} = \s $}; (2) rough diffuse {$ \mbf{I}^{\text{diff}}_\text{c} $}; (3) rough specular  {$\mbf{I}^{\text{spec}}_\text{c}$}; and (4) projector direct light image warped to the canonical camera frontal view $ \mbf{I}'_\text{p} $. 
ShadingNet consists of three subnets: (1) a surface skip branch sk1 with three convolution layers (blue blocks in the top) to pass rough ambient $ \s $ to the output layer; (2) an encoder-like rough shading branch (green blocks) that uses four convolutional layers (Conv1r-Conv4r) to downsample and extract rich features of the rough shadings; and (3) an encoder-decoder backbone network (red blocks) to capture rich direct and indirect light interactions involved in the projector-camera image formation. The backbone network consists of an encoder (Conv1-Conv4) and a decoder (Conv5, TConv1, TConv2 and Conv6).

\begin{table}[!t]
	\begin{center}
      \caption{\textbf{ShadingNet $ \F $ parameters}. \textbf{Conv} and \textbf{TConv} are convolutional and transposed convolutional layers of a backbone subnet, respectively. Note that Conv1r is the first convolutional layer of \textbf{r}ough shading branch, and Conv2r-Conv4r
      shares the same structure as backbone Conv2-Conv4. \textbf{Sk} stands for skip convolutional layers and Sk1 has three identical layers (weights are different), thus we only show one layer here. ReLU layers after each Conv/TConv/Sk layer are omitted. \textbf{Ker.}, \textbf{Str.}, \textbf{Pad.} are kernel size, stride and padding size, respectively.}\label{tab:shadingnet_param}
			{
				\begin{tabular}[]{@{}ll@{\hspace{7mm}}l@{\hspace{7mm}}c@{\hspace{3mm}}c@{\hspace{2mm}}c@{}}
					\toprule[0.5mm]
					\textbf{Layer} & \textbf{Input size}               & \textbf{Output size}               & \textbf{Ker.} & \textbf{Str.} & \textbf{Pad.} \\ \bottomrule[0.3mm]
					Conv1r         & [h, w, 9]                      & [h/2, w/2, 32]                & 3$\times$3      & 2               & 1                \\
					Conv1          & [h, w, 3]           & [h/2, w/2, 32]                & 3$\times$3      & 2               & 1                \\
					Conv2          & [h/2, w/2, 32]    & [h/4, w/4, 64]  & 3$\times$3      & 2               & 1                \\
					Conv3          & [h/4, w/4, 64]    & [h/4, w/4, 128] & 3$\times$3      & 1               & 1                \\
					Conv4          & [h/4, w/4, 128]   & [h/4, w/4, 256] & 3$\times$3      & 1               & 1                \\
					Conv5          & [h/4, w/4, 256]   & [h/4, w/4, 128] & 3$\times$3      & 1               & 1                \\
					TConv1         &[h/4, w/4, 128]   & [h/2, w/2, 64]  & 2$\times$2      & 2               & 0                \\
					TConv2         & [h/2, w/2, 64]    & [h, w, 32]          & 2$\times$2      & 2               & 0                \\
					Conv6          & [h, w, 32]         & [h, w, 3]           & 3$\times$3      & 1               & 1                \\
					Sk1            & [h, w, 32]         & [h, w, 32]          & 3$\times$3      & 1               & 1                \\
					Sk2            & [h/2, w/2, 32]               & [h/2, w/2, 32]                & 1$\times$1      & 1               & 0                \\
					Sk3            & [h/4, w/4, 64] & [h/4, w/4, 64]  & 1$\times$1      & 1               & 0                \\ \bottomrule[0.5mm]
				\end{tabular}
			}
	\end{center}
\end{table}

Firstly, in the rough shading branch, the input rough shadings ($\s, \mbf{I}^{\text{diff}}_\text{c}, \mbf{I}^{\text{spec}}_\text{c}$) are concatenated along the channel. Then, the 9-channel input is fed to the encoder (Conv1r-Conv4r) and sequentially dowmsampled to 1/4 of the original sizes, while rich features are extracted by increasing the feature map channels: $9\rightarrow32\rightarrow64\rightarrow128\rightarrow256$. Similarly, the encoder part of the backbone (Conv1-Conv4) has the same structure as the rough shading branch, except that the input is a 3-channel warped projector direct light image $ \mbf{I}'_\text{p} $. The four vertical black skip connections between the rough shading branch and the backbone indicate that rough shading branch feature maps are added to the backbone ones. Moreover, the two black horizontal skip connections under the backbone network add low-level feature maps (Conv1 and Conv2) to the \emph{first half} of the high-level feature maps (Conv5 and TConv1). Meanwhile, the two low-level feature maps of the rough shading branch (Conv1r and Conv2r) are transformed by two skip convolutional layers sk2 and sk3, and then added to the \emph{second half} of Conv5 and TConv1. The two types of skip connections allow ShadingNet to carry low-level visual details to deeper layers, and to better learn complex interactions between the direct light and the scene geometry in the feature space. Afterwards, the backbone decoder gradually upsamples the fused feature maps back to the original image sizes, meanwhile, decreases the feature map channels by $256\rightarrow128\rightarrow64\rightarrow32\rightarrow3$. Finally, the backbone network and sk1 output images are added through the blue skip connection. Detailed network ShadingNet parameters are shown in \autoref{tab:shadingnet_param}, where \textbf{Input size} and \textbf{Output size} are sizes of input and output feature map/image in the [height, width, channel] format. Note that for each layer, the number of convolutional filters is equal to the number of output channel. 

It is worth noting that the input and design of \textbf{ShadingNet} $ \F $ are particularly important to the model convergence: (1) all the inputs are geometrically aligned, obviating the need for $ \F $ to learn geometry, and (2) ShadingNet output image $  \hat{\mbf{I}}_\text{c} $ shares some structural details with the rough ambient image ($ \s $ in \autoref{fig:net}), which can be leveraged as a good initial point. Thus, we use a shallow surface skip branch to pass $ \s $ to the backbone output, allowing $ \F $ to focus on learning the residual \cite{he2016deep} light on top of $ \s $, instead of directly inferring the final result $\hat{\mbf{I}}_\text{c}$ from random guess. The two techniques and the photometric and geometric constraints below together reduce the chance of falling into local minima early on and improve convergence. 

We find that a more complex ShadingNet (\eg, a deeper network with more learnable parameters) and longer training time may increase accuracy of relighting, compensation and shape reconstruction. However, considering the practicability of DeProCams in SAR applications, in this paper we choose this architecture and 1,000 iterations to balance training/inference efficiency and accuracy.

\section{Implementation and training details}\label{sec:implementation}
We implement DeProCams $\pi$ using PyTorch \cite{paszke2017automatic} and Kornia \cite{eriba2019kornia}. The learnable parameters of DeProCams are $ \pi_{\theta} = \{\dc, \F\}_{\theta} $. In particular, the depth map $ \dc $ contains $ \text{h}\times\text{w}\times 1 $ learnable parameters, where h and w are camera image height and width, respectively; $ \F $ contains learnable convolutional filters/kernels (\autoref{tab:shadingnet_param}) and with a carefully chosen loss function, constraints and an optimizer (\eg, stochastic gradient descent and Adam \cite{kinga2015method}), both $ \F $ and $ \dc $ can be learned jointly.

\subsection{Loss function and constraints}\label{sec:constraints}
As shown in \autoref{fig:net}, given a new projector input image $ \Ip $, a well trained DeProCams $ \pi $ should be able to reconstruct/infer a relit image $ \hat{\mbf{I}}_\text{c}=\pi(\Ip)$, which is very close to the real camera-captured one $\Ic$. Thus, during network training, we encourage the similarity between DeProCams inferred and the ground truth images by minimizing the pixel-wise $L_1+\text{SSIM}$ \cite{wang2004image} loss:
\begin{equation}\label{eq:recon}
  \mathcal{L}_\text{recon} = \frac{1}{N}\sum_i^N \big(|\hat{\mbf{I}}^{(i)}_\text{c} - \Ic^{(i)}| + 1 - \text{SSIM}(\hat{\mbf{I}}^{(i)}_\text{c},\ \Ic^{(i)}) \big)
\end{equation}
where $ N $ is the number of training image pairs $ (\Ip, \Ic) $. The advantages of this image reconstruction loss are shown in \cite{zhao2017loss, huang2019compennet, huang2019compennet++, huang2020end}.

Note that directly optimizing $ \F $ and $ \dc $ using $ \mathcal{L}_\text{recon} $ may be difficult and as a result, the final depth map $ \dc $ is coarse and DeProCams inferred relit image $ \Ic $ is blurry, because $ \mathcal{L}_\text{recon} $ alone does not impose sufficient constraints. To address this issue, besides modeling rough direct light shadings and projector direct light mask in DepthToAttribute, we introduce three additional photometric and geometric constraints to ensure accurate inference of the shading attributes (\autoref{eq:attributes}) and prevent $ \F$ from overfitting. Their effectiveness is shown in ablation study \autoref{sec:ablation}.

\subsubsection{Direct light mask consistency constraint} \label{sec:mask_const}
In \autoref{sec:mask}, DeProCams explicitly infers a projector direct light mask $ \mbf{M} $ (\autoref{eq:mask}) to address occlusion. It also imposes an additional constraint that $ \mbf{M} $ should match the projector field of view (FOV) $ \s^{*}\in \mathbb{R}^{\text{h}\times\text{w}\times 1} $. This geometric constraint can reduce depth searching space by early rejecting geometrically impossible depths and is implemented as a pixel-wise $ L_2 $ loss:
\begin{equation}\label{eq:mask_loss}
\mathcal{L}_\text{mask} = \norm{\mbf{M} - \s^{*}}^2_2
\end{equation}
where $ \s^{*} $ is a thresholded camera-captured surface for projector FOV.

\subsubsection{Rough shading consistency constraint} 
Although our rough direct light shading (\autoref{sec:problem}) is different from an actual Phong model, it is useful to encourage photometric similarities between rough diffuse shading $\mbf{I}^{\text{diff}}_\text{c}$ (\autoref{eq:rough}) and camera-captured ground truth $\Ic$. We hence use the following pixel-wise $ L_2 $ loss:
\begin{equation}\label{eq:photometry_loss}
\mathcal{L}_\text{rough} =  \frac{1}{2N}\sum_i^N \norm{\mbf{I}^{\text{diff}(i)}_\text{c} - \Ic^{(i)}}^2_2
\end{equation}
This loss forces DeProCams to infer plausible light rays and surface normals. Since the two intermediate attributes are derived from the depth map $ \dc $, this constraint also improves depth map accuracy.

\subsubsection{Smoothness constraint}
In most SAR setups, the depth map $ \dc $ (\autoref{fig:net}) and intermediate attributes such as the sampling/warping grid $ \mbf{\Omega} $ (\autoref{eq:warp}) and the normal map $ \mbf{n} $ (\autoref{eq:normal}) are locally smooth. This constraint can be implemented as minimizing the absolute value of image gradient:
\begin{equation}
  \mathcal{L}_\text{smooth}(\mbf{I}) = |\partial_x\mbf{I}|e^{-||\partial_x\s||} + |\partial_y\mbf{I}|e^{-||\partial_y\s||}
\end{equation}	
where $ \mbf{I} $ is an image (\eg, $ \dc $, $ \mbf{\Omega} $ and $ \mbf{n} $) whose smoothness is constrained by its partial derivatives in $ x $ and $ y $ directions. To preserve sharp edges, we use the gradient of camera-captured scene $ \s $ as smoothness loss weights \cite{zhou2017unsupervised, godard2017unsupervised, garg2016unsupervised}. Combining smoothness losses of $ \dc $, $ \mbf{\Omega} $ and $ \mbf{n} $, we have:
\begin{equation}\label{eq:smooth_loss}
\mathcal{L}^\text{all}_\text{smooth} = 2\mathcal{L}_\text{smooth}(\dc)+\mathcal{L}_\text{smooth}(\mbf{\Omega})+0.01\mathcal{L}_\text{smooth}(\mbf{n}) 
\end{equation}	
The smoothness constraint forces $ \dc $, $ \mbf{\Omega} $ and $ \mbf{n} $ to be geometrically plausible and prevents $ \F$ from overfitting on coarse depth map and intermediate attributes in early training stages.

\subsection{Network training}
We combine image reconstruction loss $ \mathcal{L}_\text{recon} $ with the three constraints above
and jointly minimize them using Adam optimizer \cite{kinga2015method}:
\begin{align}\label{eq:opt}
  \{\dc, \F\}_\theta  =  \argmin_{\{\dc',\F'\}_\theta}\mathcal{L}_\text{recon} + \mathcal{L}_\text{mask}  + \mathcal{L}_\text{rough} + \mathcal{L}^\text{all}_\text{smooth}
\end{align}
The performance of the gradient-based optimization above depends on $ \{\dc, \F\}_\theta$ initializations and a reasonable initial point can greatly improve convergence. Thus, to initialize $ \dc $, we start by assuming the initial sampling/warping $\mbf{\Omega}$ as affine, and then we estimate the affine transformation using the projector input image size and the bounding rectangle of camera-captured projector FOV mask $ \s^{*} $ (similar to WarpingNet \cite{huang2019compennet++} affine parameters initialization). Afterwards, we compute the depths of $\mbf{\Omega}$ using triangulation \cite{hartley2003multiple} and set the initial $ \dc $ to it. ShadingNet $ \F $ filter weights are initialized using He's method \cite{he2015delving}.

The initial learning rate of the depth map $ \dc $ and the ShadingNet $ \F $ are set to $ 10^{-2} $ and $ 10^{-3} $, respectively. Then, using training image pairs like $ \{(\Ip^{(i)}, \Ic^{(i)})\}^{N}_{i=1} $ ($ N $ is \#Train in \autoref{tab:compare}), we train the model for 1,000 iterations on an Nvidia GeForce 1080Ti GPU with a batch size of 24, and it takes about 5 minutes to finish.

\begin{table*}[!t]
  \begin{center}
    \caption{\textbf{Quantitative comparison of relighting and depth reconstruction}. Results are averaged over 23 different setups. $ \mbf{d}_\text{err} $ is the mean $L_2$ distance between the inferred and the ground truth point cloud (mm). For SL \cite{moreno2012simple} without manual cleaning and interpolation, $ \mbf{d}_\text{err} = 1.0858 $. Note that SL $ \mbf{d}_\text{err}$ is much smaller because the ground truth is a manually cleaned and interpolated version of SL point cloud.
    }\label{tab:compare}
      \begin{tabular}{@{\hspace{.5mm}}l@{\hspace{5mm}}l@{\hspace{9mm}}c@{\hspace{5mm}}c@{\hspace{5mm}}c@{\hspace{7mm}}  @{\hspace{5mm}}c@{\hspace{5mm}}c@{\hspace{5mm}}c@{\hspace{9mm}}c@{\hspace{.1mm}}}
        \toprule[0.5mm]
        &  & \multicolumn{3}{@{\hspace{5mm}}c@{\hspace{18mm}}}{\textbf{SL Masked}} & \multicolumn{4}{c@{\hspace{23mm}}}{\textbf{Whole Image}} \\
        \cmidrule[0.3mm](r{12mm}){3-5} \cmidrule[0.3mm](r{9mm}){6-8}\textbf{\# Train} & \textbf{Model} & \textbf{PSNR}$\uparrow$ & \textbf{RMSE}$\downarrow$ & \textbf{SSIM}$\uparrow$ &\textbf{PSNR}$\uparrow$ & \textbf{RMSE}$\downarrow$ & \textbf{SSIM}$\uparrow$ & $\mbf{d}_\text{err} (\text{mm})\downarrow$ \\
        \midrule[0.3mm]
        50 & TPS + SL \cite{grundhofer2015robust,moreno2012simple} & 27.6703 & 0.0731 & 0.9407 & 17.8285&	0.2345&	0.4336 & - \\
        & CompenNet++ \cite{huang2019compennet++} & 24.2998 & 0.1093 & 0.8991 & 23.3141 & 0.1220 & 0.8437 & - \\
        & DeProCams (ours) & \textbf{27.9819} & \textbf{0.0719} & \textbf{0.9408} & \textbf{26.1807} & \textbf{0.0883} & \textbf{0.8919} & \textbf{8.5653} \\
        & No const. (ours) & 27.5575          & 0.0751          & 0.9365          & 25.8954          & 0.0910          & 0.8874          & 9.7733          \\
        & No mask (ours)   & 27.5614          & 0.0759          & 0.9346          & 25.8146          & 0.0927          & 0.8851          & 13.1531         \\
        & No rough (ours)  & 27.0641          & 0.0795          & 0.9313          & 25.4962          & 0.0951          & 0.8798          & 10.4029         \\
        \midrule[0.3mm]
        100 & TPS + SL \cite{grundhofer2015robust,moreno2012simple} & 27.6165 & 0.0736 & 0.9439 & 17.8261&	0.2346&	0.4365        & - \\
        & CompenNet++ \cite{huang2019compennet++} & 26.7337 & 0.0817 & 0.9329 & 25.3195 & 0.0961 & 0.8849 & - \\
        & DeProCams (ours) & \textbf{29.6620} & \textbf{0.0586} & \textbf{0.9569} & \textbf{27.3590} & \textbf{0.0768} & \textbf{0.9100} & \textbf{4.9520} \\
        & No const. (ours) & 29.1676          & 0.0621          & 0.9535          & 27.0458          & 0.0797          & 0.9064          & 6.0782          \\
        & No mask (ours)   & 29.5997          & 0.0592          & 0.9564          & 27.2395          & 0.0782          & 0.9100          & 7.2205          \\
        & No rough (ours)  & 28.8089          & 0.0649          & 0.9498          & 26.7679          & 0.0822          & 0.9005          & 5.7721          \\
        \midrule[0.3mm]
        250 & TPS + SL \cite{grundhofer2015robust,moreno2012simple} & 27.4265 & 0.0753 & 0.9444 & 17.8056 &	0.2352&	0.4369
        & - \\
        & CompenNet++ \cite{huang2019compennet++} & 26.7593 & 0.0810 & 0.9322 & 25.3926 & 0.0947 & 0.8860 & - \\
        &  DeProCams (ours) & 30.5640          & 0.0522          & 0.9655          & 28.0801          & 0.0702          & 0.9224          & \textbf{2.9408} \\
        &  No const. (ours) & 30.3678          & 0.0537          & 0.9642          & 27.9490          & 0.0714          & 0.9209          & 3.4319          \\
        &  No mask (ours)   & \textbf{30.7901} & \textbf{0.0509} & \textbf{0.9676} & \textbf{28.1284} & \textbf{0.0701} & \textbf{0.9253} & 4.1483          \\
        &  No rough (ours)  & 30.0289          & 0.0557          & 0.9603          & 27.7054          & 0.0733          & 0.9144          & 3.2261          \\
        \midrule[0.3mm]
        500 & TPS + SL \cite{grundhofer2015robust,moreno2012simple} & 27.2926 & 0.0766 & 0.9436 & 17.7908&	0.2356&	0.4362
        & - \\
        & CompenNet++ \cite{huang2019compennet++} & 27.4843 & 0.0743 & 0.9386 & 26.0587 & 0.0879 & 0.8948 & -\\
        & DeProCams (ours) & 30.8368          & 0.0506          & 0.9674          & 28.3392          & 0.0682          & 0.9251          & \textbf{2.5087} \\
        & No const. (ours) & 30.7392          & 0.0512          & 0.9666          & 28.2660          & 0.0688          & 0.9242          & 2.8683          \\
        & No mask (ours)   & \textbf{31.1380} & \textbf{0.0488} & \textbf{0.9699} & \textbf{28.4636} & \textbf{0.0675} & \textbf{0.9286} & 3.5542          \\
        & No rough (ours)  & 30.3418          & 0.0535          & 0.9626          & 27.9812          & 0.0710          & 0.9176          & 2.7140          \\
        \bottomrule[0.5mm]
      \end{tabular}
  \end{center}
\end{table*}

\subsection{One stone three birds}\label{sec:tasks}	
Once trained, DeProCams can be used for three SAR tasks: image-based relighting, projector compensation and depth/normal estimation. 

\subsubsection{Image-based relighting} 
Image-based relighting infers the camera-captured image $ \tilde{\mbf{I}}_\text{c} $ given a novel projector input lighting image $ \tilde{\mbf{I}}_\text{p} $ without actual projection or capture. Since DeProCams $ \pi $ has already learned the forward projector-camera image mappings (\autoref{eq:render_noncalib}), relighting is performed by a single forward pass of DeProCams by $\tilde{\mbf{I}}_\text{c} = \pi(\tilde{\mbf{I}}_\text{p})$.

\subsubsection{Projector compensation}
Given a new projector input image $ \Ip $, projector compensation aims to infer a compensation image $ \mbf{I}^{*}_\text{p} $, such that when projected to the textured surface, the camera-captured image is ideally the same as $ \Ip $ (\ie, without geometric or photometric disturbances from the surface). In practice, the optimal displayable area of the projection surface is restricted to a subregion of the projector FOV, and following \cite{huang2019compennet++} we first compute an optimal desired camera-captured image $ \Ic' $,  which is a scaled version of $ \Ip $. Then, projection compensation is to find $ \mbf{I}^{*}_\text{p} $ such that $ \Ic' = \pi(\mbf{I}^{*}_\text{p}) $ and it can be solved using the following objective function:
\begin{align}\label{eq:cmp}
\mbf{I}^{*}_\text{p}  =  \argmin_{\Ip}\big(\underbrace{\abs{\pi(\Ip) - \Ic'}+1-\text{SSIM}(\pi(\Ip), \Ic')}_{\text{image reconstruction loss}} + \nonumber\\ 
\underbrace{\mathcal{L}_\text{smooth}(\Ip)}_{\text{smoothness loss}} + 
\underbrace{10\left(\left\|\max(\Ip-1, 0)\right\|^2 + \left\| \min(\Ip, 0)\right\|^2\right)}_{\text{color saturation loss \cite{grundhofer2015robust}}}\big)
\end{align}

\subsubsection{Depth and normal}
The depth $ \dc $ is inherently learned by DeProCams. To be robust against setup scale variances, we normalize both $ \dc $ and the projector-camera translation vector $ \mbf{t} $ by dividing $ \Vert\mbf{t}\Vert $, the actual scene depth is given by denormalization: $ \hat{\dc} = \dc*\Vert\mbf{t}\Vert $. Moreover, following the convention of learning-based depth estimation \cite{zhou2017unsupervised}, we estimate the inverse depth $ 1/ \dc $ instead to bound the learnable parameter values to $ [0, 1] $. The normal is then given by \autoref{eq:normal}.

\section{Experimental Evaluations}\label{sec:experiments}
\subsection{System configuration}
Our setup consists of a single Canon 6D camera and a single ViewSonic PJD7828HDL DLP projector, and they are geometrically calibrated using \cite{huang2020flexible}. Their resolutions are 320$\times$240 and 800$\times$600, respectively. The distance between the projector-camera system and the target scene is around 1,000mm to 2,000mm, depending on different setups. To improve data capturing efficiency, we set the camera to live view mode and connect an Elgato Cam Link 4K video capture card to the camera HDMI output. Considering the the video capture card delay and the projector-camera synchronization, we manually set a 200ms delay between the projection and the capture operation, and in total it takes about 360ms to capture a frame.

\subsection{Evaluation benchmark}
As mentioned in \autoref{sec:tasks} that DeProCams can perform three tasks, \ie, image-based relighting, projector compensation and depth/normal estimation. 
In this section, we quantitatively and qualitatively evaluate and compare DeProCams with other methods on the three tasks.

Following \cite{huang2019compennet++}, we captured 23 different setups with various lightings, poses, focal lenses and projector settings. Using the sampling patterns in \cite{huang2019compennet++}, for each setup, we captured a surface image $\s$ and $N$ (\#Train) training image pairs $ \{(\Ip^{(i)}, \Ic^{(i)})\}^{N}_{i=1} $ (took about 3 min) for training and another 200 image pairs for testing. Note that depth and normal are learned by the network without their ground truth images. 

We then captured projector compensation setups in complex scenes with glasses and slight occlusions (see \autoref{fig:exp_compensation}) to show that the proposed DeProCams outperforms the other methods. However, these scenarios may cause missing pixels and thus it is unfair to use the surrogate protocol in \cite{huang2019compennet++}. Instead, we compared on actual camera-captured compensations. Afterwards, we captured the ground truth point cloud using Gray-coded SL \cite{moreno2012simple}, followed by manual cleaning and interpolation. Future works on projector-camera image-based relighting and depth/normal reconstruction can be evaluated and compared with DeProCams without actual projections.

\begin{figure*}[!t]
  \begin{center}
    \includegraphics[width=1\linewidth]{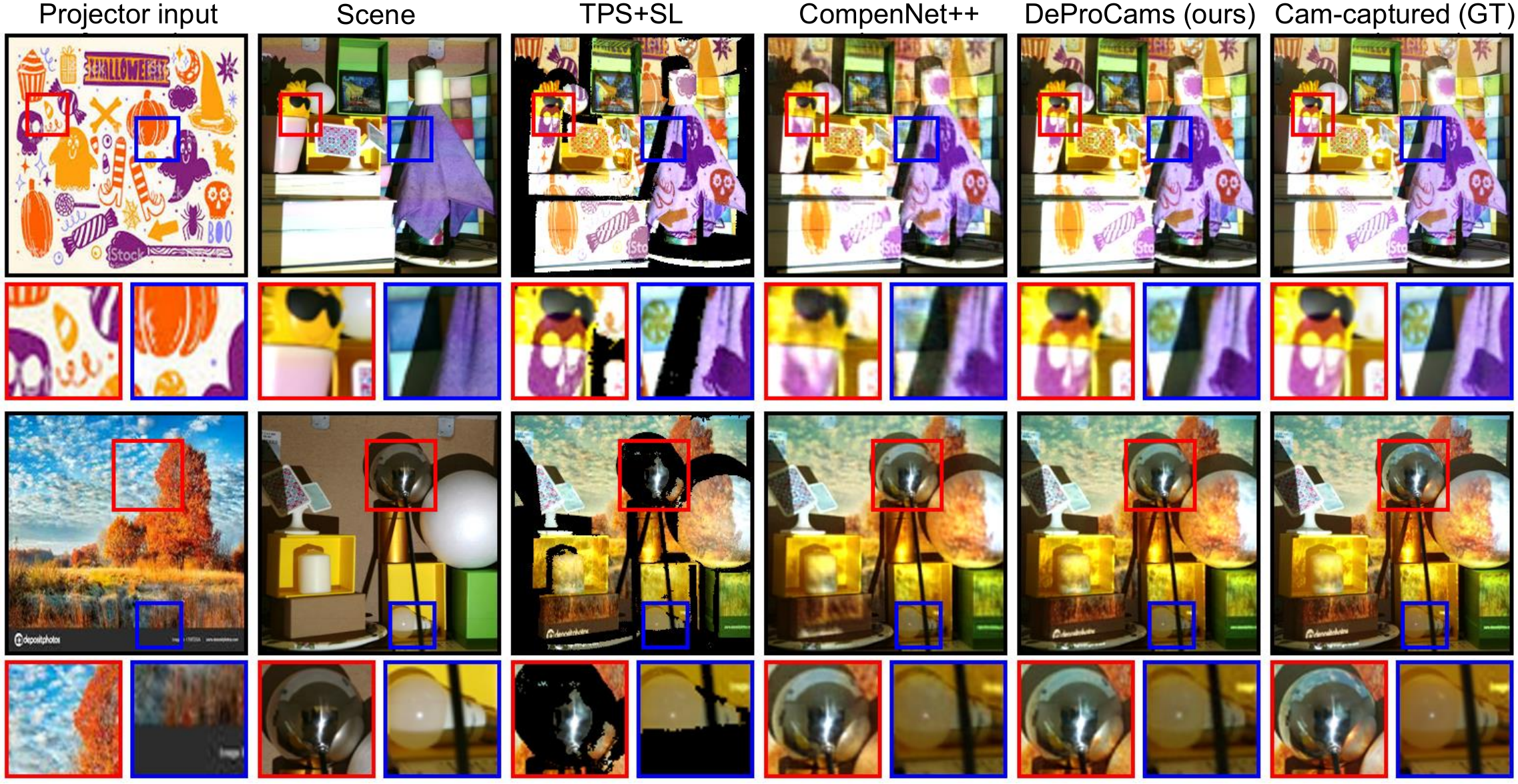}
    \caption{\textbf{Image-based relighting}. We provide two scenes (the first two rows and the last two rows) and each scene is under different projector lightings. Each image is provided with two zoomed-in patches for detailed comparison. See supplementary material for more results.}\label{fig:exp_relighting}
  \end{center}
\end{figure*}

\subsection{Image-based relighting}
Existing LTM-based relighting methods usually rely on radiometric calibrations and additional optical devices, such as lasers and a second camera \cite{wang2009kernel} and beamsplitters \cite{o2010optical}, thus we cannot compare with them using our radiometrically uncalibrated single projector-camera setup. Instead, we modified the recently proposed learning-based projector compensation method CompenNet++ \cite{huang2019compennet++} by swapping its input and output, such that forward projector-camera image mappings can be learned. We train it with the same batch size of 24 and 1,000 iterations as DeProCams, and it also takes about 5 minutes.
	
Another method that works for radiometrically uncalibrated single projector-camera setup is Grundh{\"o}fer and Iwai \cite{grundhofer2015robust} that applies pixel-wise thin-plate-spine (TPS) to projector compensation, we combine a Gray-coded SL \cite{moreno2012simple} geometric correction with it and name the method \textbf{TPS+SL}. Since SL can only recover direct light pixels as shown in \autoref{fig:exp_relighting}, for fair comparison, we compare with TPS+SL only in SL masked regions and name it \textbf{SL Masked} in \autoref{tab:compare}. DeProCams has a similar performance as TPS+SL when using 50 training images. However, TPS+SL uses 42 extra SL images to find pixel mappings. Moreover, TPS+SL cannot estimate indirect light and thus has worse performance in \autoref{tab:compare} \textbf{Whole Image} columns. 

The qualitative comparisons are shown in \autoref{fig:exp_relighting}. The first two columns are projector input light patterns and camera-captured scenes, respectively. The 3\textsuperscript{rd} to 5\textsuperscript{th} columns are relit results of different methods. The last column is camera-captured ground truth, \ie, the  1\textsuperscript{st} column projected to the  2\textsuperscript{nd}. The setups consist of challenging direct and indirect light effects, \eg, in the 1\textsuperscript{st} row, we present diffuse (plywood board, paper boxes and books), translucency (light bulb), subsurface scattering (wax, towel), specular highlight (mug) and shadow; and in the 2\textsuperscript{nd} row, we show diffuse (plywood board, paper boxes, foam ball), subsurface scattering (wax), mirror (metal ball), specular highlight (metal can), translucency (light bulb) and shadow.

For TPS+SL \cite{grundhofer2015robust,moreno2012simple}, indirect light regions such as shadow, specular highlight and mirror are black because SL \cite{moreno2012simple} rely on direct light and is unable to find correct pixel mappings in those regions (see zoomed-in patches). Moreover, it fails to estimate image blur, thus the images look much sharper than the ground truth. CompenNet++ \cite{huang2019compennet++} can estimate some indirect light but it generates much blurry relit results, \eg, the purple skull in the 1\textsuperscript{st} row red zoomed-in patch and the yellow wheel in the blue zoomed-in patch. By contrast, our DeProCams can estimate more accurate indirect light, \eg, in the 2\textsuperscript{rd} row red zoomed-in patch, the bottom-right corner of the metal ball shows reflected orange color from the white foam ball. See supplementary material for more results.

The results in \autoref{fig:exp_relighting} and \autoref{tab:compare} show that the proposed DeProCams significantly outperforms CompenNet++ \cite{huang2019compennet++} because (1) DeProCams explicitly models light-geometry-material interactions using rough direct light (\autoref{eq:rough}) and it imposes additional photometric and geometric constraints (\autoref{sec:constraints}). (2) Without depth, CompenNet++'s 2-D geometric correction is unable to address occlusions, while DeProCams predicts a differentiable projector direct light mask to explicitly deal with occlusions (\autoref{sec:mask}). (3) DeProCams's ShadingNet $\F$ refines rough shadings and has a novel structure that allows modeling more complex light-attribute interactions (\autoref{fig:net}).

\begin{table}[!t]
  \begin{center}
    \caption{\textbf{Quantitative comparison of projector compensation}. Results are averaged over 10 different setups.}\label{tab:compare_cmp}
    \begin{tabular}{l@{\hspace{9mm}}r@{\hspace{7mm}}c@{\hspace{7mm}}c}
      \toprule[0.5mm]
      \textbf{Model} & \textbf{PSNR}$\uparrow$ & \textbf{RMSE}$\downarrow$ & \textbf{SSIM}$\uparrow$ \\
      \midrule[0.3mm]
      TPS + SL \cite{grundhofer2015robust,moreno2012simple} & 21.3601&	0.1541&	0.8695\\
      CompenNet++ \cite{huang2019compennet++} & 22.2046&	0.1421&	0.8838 \\
      DeProCams (ours) & \textbf{23.8245}&	\textbf{0.1132}&	\textbf{0.8885}\\
			Uncompensated & ~9.3374 & 0.5979 &	0.1089\\			
      \bottomrule[0.5mm]
    \end{tabular}
  \end{center}
\end{table}

\subsection{Projector compensation}
We compare DeProCams with TPS+SL \cite{grundhofer2015robust,moreno2012simple} and CompenNet++ \cite{huang2019compennet++} using the same sampling image pairs as image-based relighting, except that both TPS+SL's and CompenNet++'s input and output are swapped to learn the backward mapping $ \Ic \mapsto \Ip $. Then, we train CompenNet++ for 1,500 iterations to match the original settings in \cite{huang2019compennet++}. 

The quantitative results are shown in \autoref{tab:compare_cmp}. 
Note that DeProCams outperforms both TPS+SL and CompenNet++, and has much smaller pixel-wise color errors (PSNR/RMSE). Note that we capture another 10 different setups for compensation with slight or no occlusions.

\begin{figure*}[!t]
  \begin{center}
    \includegraphics[width=1\linewidth]{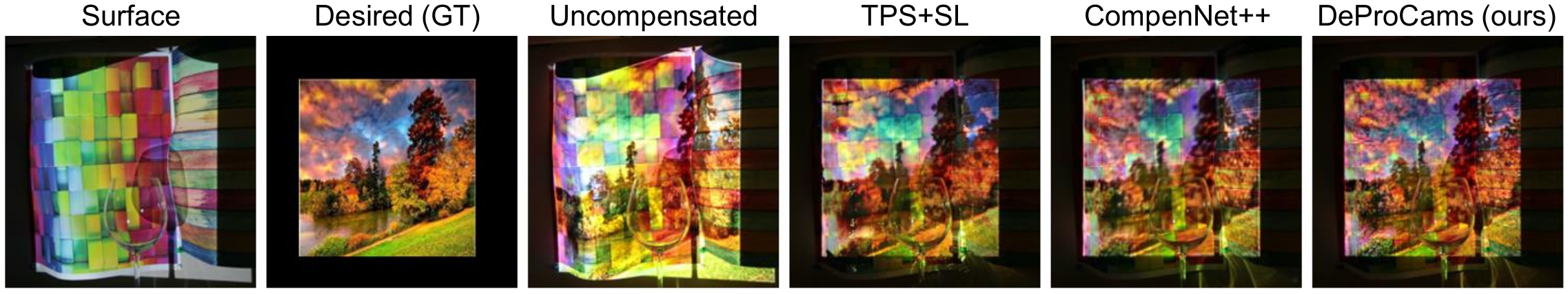}
    \caption{\textbf{Projector compensation}. The 1\textsuperscript{st} to 3\textsuperscript{rd} columns are the projection surface, optimal desired viewer perceived image and uncompensated projection, respectively.	The rest columns are real camera-captured results of different compensation methods. See supplementary for more results. 
			}\label{fig:exp_compensation}
  \end{center}
\end{figure*}

The qualitative comparisons are shown in \autoref{fig:exp_compensation}, where a challenging wine glass is placed in front of the textured surface, and some projector pixels are occluded by the wine glass stem and the bottom of the textured surface. 
Note that DeProCams generates better compensation in the sky region, while the other two methods cannot well compensate the surface texture, and thus we can see clear green cube-like artifacts. Moreover, CompenNet++ \cite{huang2019compennet++}'s result looks blurry and contains clear sawtooth patterns in the top, because it assumes a smooth projection surface and WarpingNet may fail to learn accurate pixel mappings around occluded regions. On the other hand, TPS+SL can handle occlusions, but in the top-left region some SL decoding errors deteriorate the compensation effect. See supplementary material for more results.

In summary, DeProCams outperforms TPS+SL \cite{grundhofer2015robust,moreno2012simple} and CompenNet++ \cite{huang2019compennet++} on this projector compensation benchmark because (1) geometry- and attributes-aware formulation (\autoref{sec:problem}) and the novel architecture (\autoref{fig:net}) let DeProCams better model complex projector-camera image mappings. (2) Occlusion is explicitly addressed by a differentiable projector direct light mask.
(3) We explicitly deal with saturation errors due to projector's physical limitation and encourages compensation image smoothness (\autoref{eq:cmp}). (4) The fully differentiable architecture allows the combined photometric and geometric constraints to optimize the compensation image through backpropagation.

\begin{figure}[!t]
  \begin{center}
    \includegraphics[width=1\linewidth]{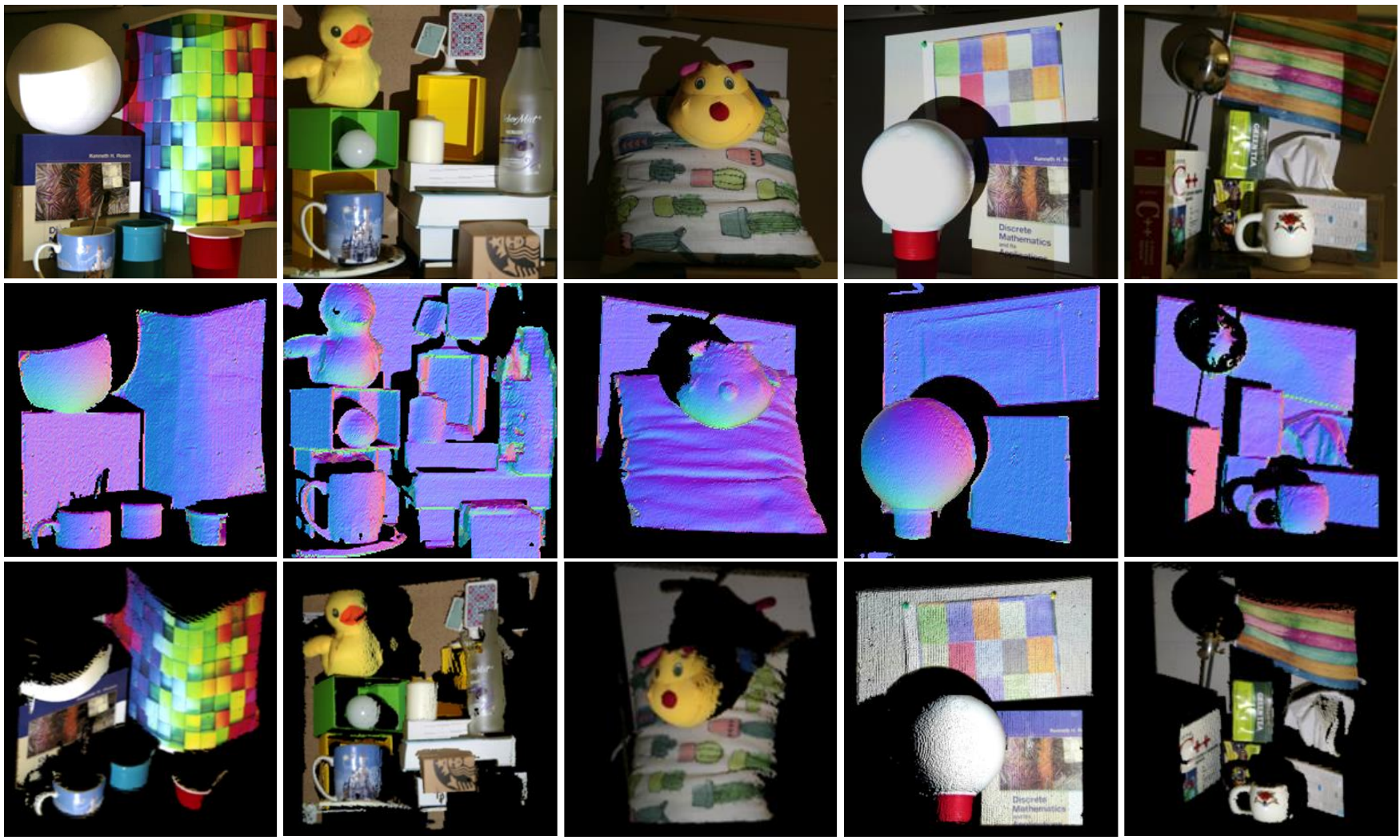}
    \caption{\textbf{Depth and normal reconstruction}. We show five different setups. The {1\textsuperscript{st} row} is the scene under plain illumination.  The {2\textsuperscript{nd} row} is estimated normal. The {3\textsuperscript{rd} row} is reconstructed point cloud viewed from a different pose.  See supplementary material for more results. }\label{fig:exp_reconstruction}
  \end{center}
\end{figure}

\subsection{Depth and normal estimation}
To evaluate shape reconstruction, we capture the ground truth point cloud using Gray-coded SL \cite{moreno2012simple}, followed by manual cleaning and interpolation. 
Then, we convert DeProCams learned depth map $\hat{\dc}$ to a point cloud using \autoref{eq:coord_map} and calculate depth error $ \mbf{d}_\text{err} $ as the mean Euclidean distance between the estimated and ground truth point cloud. We also compare with the SL's raw point cloud before it is manually cleaned and interpolated. Note this comparison gives SL \cite{moreno2012simple} great advantages, thus its depth error $ \mbf{d}_\text{err} = 1.0858 $ mm is much smaller than those of DeProCams. 

The quantitative and qualitative results are shown in \autoref{fig:exp_reconstruction} and \autoref{tab:compare} and the proposed DeProCams shows good depth and normal reconstruction capability. Moreover, unlike SL that assumes no geometric or photometric constraints between the surfaces points, DeProCams uses local weighted smoothness and rough shading consistency to generate high fidelity shapes, Thus, as shown in \autoref{fig:exp_ablation}, our point cloud and normal map have lighter aliasing artifacts than SL.

\subsection{Ablation study}\label{sec:ablation}
To show the effectiveness of the differentiable projector direct light mask (\autoref{sec:mask}) and the photometric and geometric constraints (\autoref{sec:constraints}), we performed comprehensive ablation studies. We created degraded DeProCams with the respective constraint disabled, \ie, \textbf{No rough}, \textbf{No mask} and \textbf{No const.} are w/o rough diffuse and specular shadings (\autoref{eq:rough}),  w/o direct light mask, and w/o rough shading consistency (\autoref{sec:constraints}), respectively. 

\begin{figure}[!t]
  \begin{center}
    \includegraphics[width=1\linewidth]{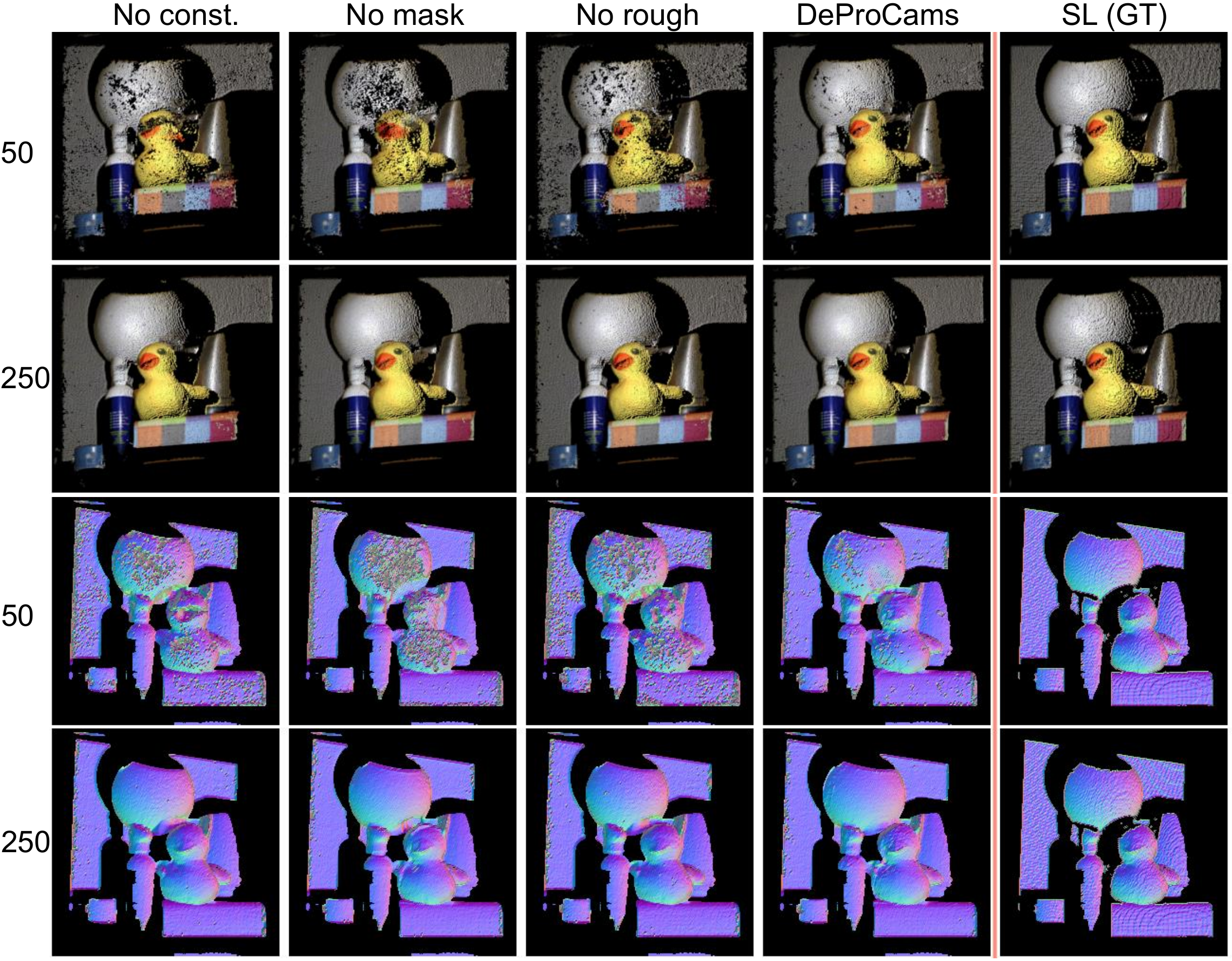}
    \caption{\textbf{Ablation study of DeProCams constraints}. The first two rows are reconstructed point cloud viewed from a new pose and the last two rows are reconstructed normal. See supplementary for more results. }\label{fig:exp_ablation}
  \end{center}
\end{figure}

By comparing DeProCams and its degraded versions in \autoref{tab:compare}, we find that DeProCams obtained the best relighting results when \#Train$<= 100$ and the best depth reconstruction results ($\mbf{d}_\text{err}$) for all \#Train. To visually interpret $\mbf{d}_\text{err}$, we show the reconstructed point cloud and normal in \autoref{fig:exp_ablation}. For each section, we compare different number of training images, \ie, \#Train = 50 and 250. We repeat the ground truth of manually cleaned and interpolated SL depth and normal (named \textbf{SL (GT)}) for each row for convenient visual comparison.
\textbf{When \#Train is large}, the quantitative and qualitative results of DeProCams and its degraded versions look similar, indicating that the hidden light and geometry interactions and constraints may also be learned without explicit modeling, as long as sufficient training data is provided. However, unlike general image-to-image translation tasks that data can be collected offline, SAR applications require online data capturing and thus are restricted to smaller training datasets. In this case, the proposed light and geometry interaction and constraints become particularly important. 
\textbf{When \#Train is small}, DeProCams point cloud and normal are smoother than its degraded versions, showing that explicitly modeling rough light and geometry interactions (even different from an empirical Phong model) and adding photometric and geometric constraints can improve model convergence when training samples are limited. See supplementary material for more setups.

\section{Discussion}\label{sec:discussion}
\noindent\textbf{Applicability to other settings.} 
The proposed DeProCams can be applied to higher resolution projectors and cameras without modifying the network structure and filter sizes, as long as the machine has sufficient GPU memory and computing power. Theoretically, a higher resolution projector-camera pair can provide finer training image details, which may improve DeProCams performance quality. Note that for new resolutions sizes not divisible by 4, we can change the padding size in \autoref{tab:shadingnet_param} to compensate the input/output feature map sizes.

For miniaturized projector-camera systems with limited computing power, we can use model compression techniques \cite{iandola2016squeezenet} and half-precision floating-point to create a lighter version of DeProCams to improve efficiency with some sacrifice in quality. One concern about miniaturized projector-camera systems is limited dynamic ranges, which may cause dropped DeProCams performance.

\noindent\textbf{Limitations and future work.} 
DeProCams is designed for SAR applications and has a different projector-camera setup from LTM, and its image-based relighting in indirect light regions still have room for improvement, as shown in \autoref{fig:exp_relighting}, the 1\textsuperscript{st} row wax and light bulb and \autoref{fig:exp_reconstruction} and the 2\textsuperscript{nd} column frosted glass bottle. Incorporating domain knowledge of LTM and adding an additional step of radiometric calibration may improve the results and it is definitely an interesting direction to explore in future work.

\section{Conclusions}\label{sec:conclusions}
In this paper, we propose the first end-to-end trainable solution, DeProCams, to solve the three important SAR tasks: image-based relighting, projector compensation and depth/normal reconstruction simultaneously. In addition, by solving the three tasks in a unified model, DeProCams waives the need for potential additional optical devices, radiometric calibrations and structured light patterns.
To improve convergence and performance, especially when sampling images are limited, we leverage rich photometric and geometric constraints such as projector direct light mask, rough direct light shading and smoothness constraint. In our thorough experiments, DeProCams shows clear advantages over previous approaches with promising quality.
Moreover, by constructing the first simultaneous image-based relighting and depth/normal estimation benchmark for projector-camera systems, our work is expected to facilitate future SAR studies in this direction.

\noindent\textbf{Acknowledgments.} This work was supported in part by US National Science Foundation (No. 2006665 and 1814745).

\bibliographystyle{abbrv-doi}

\bibliography{ref}
\end{document}